# A Self-Supervised Transformer for Unusable Shared Bike Detection

Yin Huang [#]
*School of Transportation and Logistics*
*Southwest Jiaotong University*
Chengdu, China
huangyin@my.swjtu.edu.cn

Yongqi Dong [#, *]
*Chair and Institute of Highway Engineering*
*RWTH Aachen University*
Aachen, Germany
yongqi.dong@rwth-aachen.de

Youhua Tang [*]
*School of Transportation and Logistics*
*Southwest Jiaotong University*
Chengdu, China
tyhctt@swjtu.cn

Alvaro García Hernandez
*Chair and Institute of Highway Engineering*
*RWTH Aachen University*
Aachen, Germany
alvaro@isac.rwth-aachen.de

[#] These authors contributed equally and are co-first authors. [*] Corresponding authors (tyhctt@swjtu.cn, yongqi.dong@rwth-aachen.de).

*Abstract*—The rapid expansion of bike-sharing systems (BSS) has greatly improved urban "last-mile" connectivity, yet large-scale deployments face escalating operational challenges, particularly in detecting faulty bikes. Existing detection approaches either rely on static model-based thresholds that overlook dynamic spatiotemporal (ST) usage patterns or employ supervised learning methods that struggle with label scarcity and class imbalance. To address these limitations, this paper proposes a novel Self-Supervised Transformer (SSTransformer) framework for automatically detecting unusable shared bikes, leveraging ST features extracted from GPS trajectories and trip records. The model incorporates a self-supervised pretraining strategy to enhance its feature extraction capabilities, followed by fine-tuning for efficient status recognition. In the pretraining phase, the Transformer encoder learns generalized representations of bike movement via a self-supervised objective; in the fine-tuning phase, the encoder is adapted to a downstream binary classification task. Comprehensive experiments on a real-world dataset of 10,730 bikes (1,870 unusable, 8,860 normal) from Chengdu, China, demonstrate that SSTransformer significantly outperforms traditional machine learning, ensemble learning, and deep learning baselines, achieving the best accuracy (97.81%), precision (0.8889), and F1-score (0.9358). This work highlights the effectiveness of self-supervised Transformer on ST data for capturing complex anomalies in BSS, paving the way toward more reliable and scalable maintenance solutions for shared mobility.

*Keywords—bike-sharing systems, fault detection, Transformer, self-supervised learning, spatiotemporal feature extraction, GPS trajectory analysis*

## I. Introduction

Bike-sharing systems (BSS) have witnessed exponential growth globally, emerging as a sustainable and convenient solution for short-distance urban transportation. This innovative system not only addresses last-mile connectivity challenges in urban mobility but also effectively reduces carbon emissions, mitigates traffic congestion, and significantly enhances travel convenience and satisfaction for passengers [1], [2]. However, the large-scale deployment of BSS has introduced critical operational challenges, particularly in managing mechanical failures that render bikes unusable. These unusable faulty units not only inefficiently occupy public infrastructure resources but also present significant safety risks to users [3], [4]. Moreover, the accumulation of malfunctioning bikes reduces system reliability and drives up maintenance costs, necessitating robust automated detection solutions for sustainable operations.

Existing fault detection methods in BSS remain limited and primarily rely on two methodological paradigms: Model-based approaches and data-driven machine learning (ML) approaches. Model-based approaches typically involve manual feature extraction, coupled with statistical modeling to estimate the Probability of Unusability (PoU). To give an example, Kaspi et al. developed a Bayesian framework that integrates trip transaction data, user preferences, and station idle times to predict real-time PoU and quantify defective bike counts at the station level [5]. Similarly, Pal et al. leveraged a Poisson regression model to analyze correlations between bike failure rates and usage patterns (e.g., travel distance, duration, and unlock frequency), enabling the identification of malfunctioning bikes [6]. Furthermore, some studies have incorporated incentive-based mechanisms into model designs to improve fault detection accuracy and optimize bike redistribution through user feedback [7]–[9]. While these approaches leverage historical data and user feedback, their reliance on static thresholds fails to capture the dynamic spatiotemporal (ST) variations inherent in BSS and urban mobility patterns, consequently motivating the development of adaptive data-driven detection techniques.

Data-driven approaches for faulty bike detection primarily employ either unsupervised or supervised ML techniques. In the unsupervised ML paradigm, Delassus et al. developed a K-means clustering framework that processes Citi Bike's open data through real-time feature extraction and anomaly detection algorithms to identify potentially faulty bikes [10]. Separately, Zhou et al. proposed a functional principal component analysis (FPCA) approach that systematically evaluates bike availability patterns by analyzing anomalous trip data characteristics [11]. These unsupervised methods eliminate dependence on labeled training data by leveraging latent patterns. However, their reliance on intrinsic data patterns often makes it difficult to reliably differentiate mechanical failures from normal usage anomalies, reducing their practical utility and prompting a shift toward supervised approaches. In the supervised

This work was supported by the German Federal Ministry for Digital and Transport (BMDV) under the mFUND project: HarMobi.

paradigm, Alhussam et al. proposed a Hidden Markov Model (HMM) framework to predict latent operational states in BSS, analyzing correlations between cycling patterns and mechanical states to reveal novel behavioral insights into user interactions with malfunctioning bikes [12]. Additionally, Zhou et al. introduced a hybrid Q-learning-PageRank algorithm that dynamically identifies unavailable bikes and ranks functional unit availability [13]. These methods leverage labeled datasets to achieve higher accuracy, but they face scalability challenges when applied to metropolitan-scale BSS deployments and are hindered by severe class imbalance, a common issue in real-world BSS data where faulty bikes are significantly outnumbered by operational ones. These shortcomings underscore the urgent need for frameworks capable of addressing class imbalance, data scarcity in faulty bikes, and the spatiotemporal complexity in BSS data and urban mobility.

Furthermore, in the recent decade, the rapid growth of artificial intelligence, and especially the advent of deep learning (DL), has transformed fault detection in transportation systems by enabling autonomous feature extraction and the modeling of complex ST patterns, e.g., in [14]–[16]. Unlike traditional methods, DL frameworks, such as convolutional neural networks (CNNs), Long Short-Term Memory (LSTM) neural networks, and particularly Transformers, leverage hierarchical architectures to automatically derive discriminative representations from raw data, reducing reliance on domain-specific expertise. These architectures excel at capturing intricate patterns through multilayered nonlinear transformations, allowing the detection of subtle anomalies that evade conventional threshold-based approaches. Furthermore, DL models can enhance robustness by integrating multimodal data sources, such as time-series sensor readings (e.g., GPS trajectories) and trip records, into unified latent representations, mitigating the fragmentation inherent in unimodal analyses [17], [18]. This capability makes DL particularly well-suited for BSS fault detection, where diverse data streams offer complementary insights into bike operational status.

With these insights to address the identified gaps in current research, this study proposes a self-supervised Transformer model tailored for detecting unusable shared bikes. Leveraging real-world GPS trajectories and trip data from Chengdu, China, in September 2021, this paper compared the key ST characteristics of faulty and normal bikes. A self-supervised Transformer model (SSTransformer) was developed, which enhances its feature extraction capabilities through pretraining. Subsequently, by inheriting and fine-tuning the pre-trained feature encoder, SSTransformer can efficiently recognize and detect the status of shared bikes (i.e., normal or unusable). This study compared the performance of the SSTransformer with various baseline models, including traditional ML algorithms (e.g., Decision Tree and Support Vector Machine), ensemble ML methods (e.g., Random Forest and eXtreme Gradient Boosting), and other DL models (e.g., Gated Recurrent Unit model, LSTM, and traditional Transformer). The experimental results demonstrate that the proposed SSTransformer outperforms the current state-of-the-art benchmarks in the detection of unusable shared bikes, showcasing its superior effectiveness and practicality in this domain.

In short, the main contributions of this paper lie in:
- Empirical analysis of GPS trajectory and trip order patterns to uncover discriminative ST signatures for faulty and operational bikes;
- Development of an innovative self-supervised Transformer architecture optimized for ST feature extraction in fault detection;
- A self-supervised pretraining scheme that enables the Transformer encoder to learn generalized movement representations without manual labels;
- Fine-tuning of the pre-trained Transformer encoder on a downstream classification task, achieving state-of-the-art performance on a real-world dataset;
- Extensive comparisons against twelve baselines, spanning traditional ML, ensemble ML, and DL, demonstrating the proposed framework's superior accuracy, robustness to class imbalance, and scalability.

II. DATA ANALYSIS AND FEATURE ENGINEERING

This study utilized historical Origin-Destination (OD) data and GPS trajectories of shared bikes in Chengdu City, Sichuan Province, China, spanning September 11-13, 2021. The dataset comprised 10,730 bikes, including 8,860 operational bikes and 1,870 malfunctioning (unusable) bikes. The raw data provides fields such as bike ID, riding time, start and end latitude and longitude coordinates for each trip, and GPS trajectories in latitude and longitude. From these fields, five important ST features were derived: GPS trajectory coordinates (latitude and longitude), 3-day cumulative riding distance, trip frequency, and total travel time. These features capture both static attributes (e.g., prolonged idleness) and dynamic behavioral patterns (e.g., movement trajectories, frequent short-term usage) of bikes. This integration of static and dynamic features enables a comprehensive multidimensional analysis of characteristics related to normal and unusable bikes.

The subsequent analysis systematically examines disparities between normal operational and malfunctioning unusable bikes across four dimensions: usage frequency, spatial distribution patterns, historical trajectory behaviors, and feature dimensionality reduction.

A. Cycling Frequency Distribution

Fig. 1 illustrates the usage frequency distribution of normal and unusable shared bikes over a three-day period, revealing a clear divergence between the two groups. As expected, unusable bikes generally exhibited lower usage frequencies than their functional counterparts. This trend indicates that some bikes displayed signs of deterioration prior to being officially designated as malfunctioning, leading to reduced utilization. However, a distinct subset of malfunctioning bikes recorded relatively high usage frequencies, with peak instances reaching up to 20 times. This implies that certain malfunctions may occur abruptly, making it difficult to distinguish these faulty bikes from functional ones based solely on historical usage data.

Consequently, while historical usage frequency may serve as a preliminary indicator for malfunction detection, relying exclusively on it alone could yield incomplete or unreliable diagnostics.

*B. Spatial Distribution*

Fig. 2 depicts the spatial hotspot distribution of normal and unusable shared bikes in Chengdu. Hotspots were generated using the Folium map API, based on the normalized quantity of each bike group, to visualize potential spatial correlations with the rates of malfunctioning.

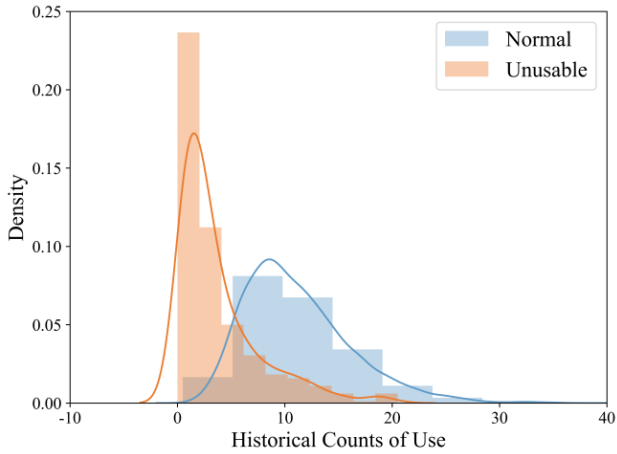

Fig. 1. The distribution of shared bike usage frequency.

A pronounced spatial divergence is evident: operational bikes demonstrate extensive coverage, spanning from the urban core to peripheral regions, consistent with current deployment strategies. In contrast, malfunctioning unusable bikes exhibit minimal clustering in the city center and are predominantly concentrated in the second and third-ring areas. This distribution may reflect heightened maintenance efficiency in the central business district, where resources are more readily available, whereas peripheral zones, characterized by dispersed deployment and lower usage intensity, experience elevated malfunction rates.

*C. Spatio-temporal Trajectory*

Dynamic usage patterns play a pivotal role in evaluating the operational status of shared bikes. Fig. 3 compares the trajectories of one typical normal operational bike (a) and one typical unusable malfunctioning bike (b).

Typical trajectories of operational bikes, as in Fig. 3 (a), are smooth and continuous, indicating uninterrupted operation. In contrast, trajectories of malfunctioning bikes, as in Fig. 3 (b), show dense point clusters and repeated short-distance trips near destination zones, suggesting mechanical issues that force premature trip termination. Such behavioral signatures, i.e., trajectory fragmentation and clustering, provide valuable cues for faulty bike detection beyond simple trip counts or locations, as well as valuable insights for maintenance strategies in BSS.

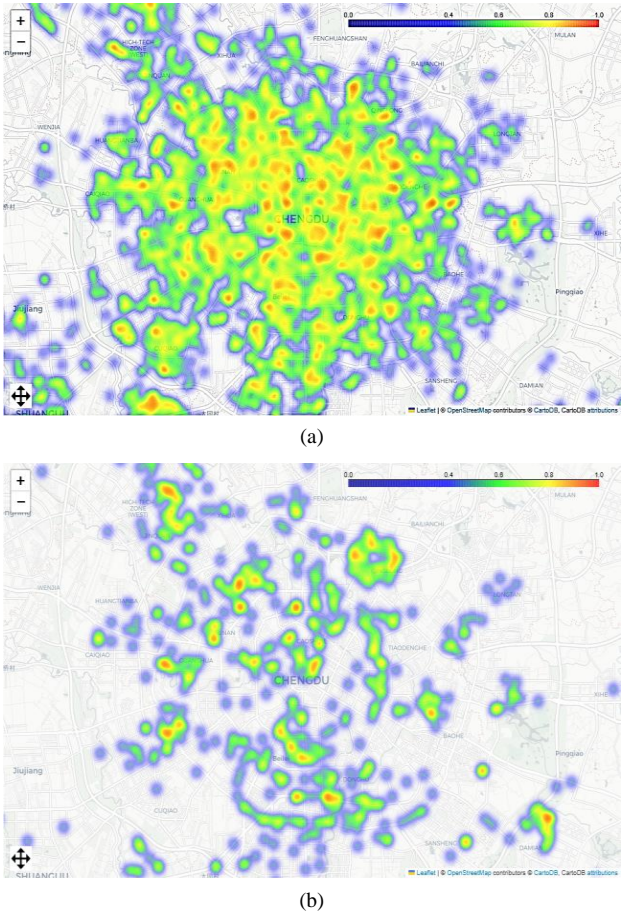

Fig. 2. Spatial hotspot distribution of (a) normal and (b) unusable bikes

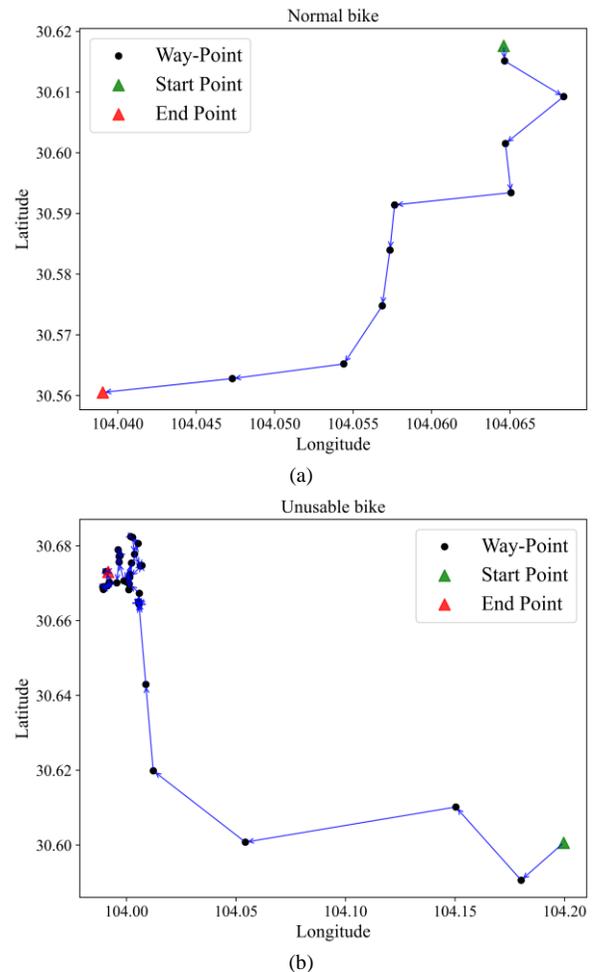

Fig. 3. Typical trajectories of shared bikes: (a) normal and (b) unusable

## D. T-SNE Dimensionality Reduction

To comprehensively evaluate the combined influence of the five selected ST features (i.e., latitude and longitude in trajectories, cumulative riding distance, trip frequency, and total travel time), t-SNE dimensionality reduction was applied to project these multidimensional features into a two-dimensional space. Fig. 4 reveals distinct clustering patterns for normal operational and unusable malfunctioning bikes, affirming the feature set's effectiveness in identifying commonalities among faulty units. Nevertheless, the presence of overlapping clusters between the two groups highlights the necessity for an optimized model to make use of these features for enhanced classification accuracy.

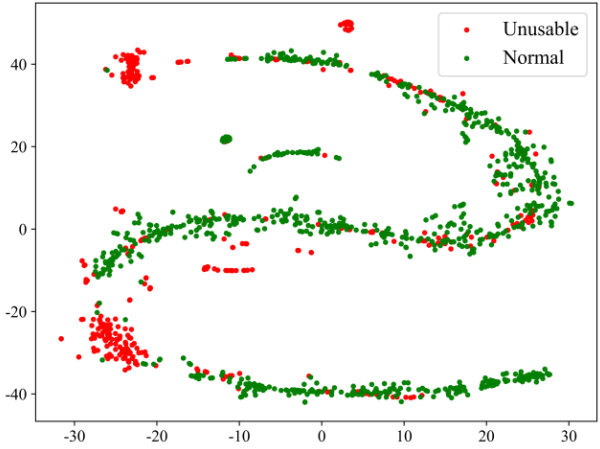

Fig. 4. T-SNE dimensionality reduction visualization of selected features

## III. METHODOLOGY

### A. Problem Formulation

The data-driven unusable bike detection task can be described as follows. Given raw data $X = [x_1, x_2, ..., x_N] \in \mathbb{R}^{N \times T \times D}$ ($N$: bike count, $T$: time steps, $D$: feature dimensions) from BSS, the goal is to determine a mapping function that projects $X$ onto a target matrix $Y = [y_1, y_2, ..., y_N]$. Formally, the approach can be represented as:

$$Y = \phi(X; \Theta) \quad (1)$$

where $\phi$ is the feature mapping function, $\Theta$ denotes the collection of weight parameters in the function. The status of the $i$-th bike is represented by a binary indicator:

$$y_i = \begin{cases} 0 & \text{(Normal)}, \\ 1 & \text{(Unusable)} \end{cases} \quad (2)$$

### B. Overall Solution Pipeline

This study addresses the unusable bike detection as a binary class prediction problem, leveraging Transformer as the DL model with ST features as input data. To enhance the Transformer's capability in extracting discriminative ST patterns, this study introduces a pretraining phase with reconstructing masked features as the task. Fig. 5 delineates the overall solution pipeline, which systematically integrates data consolidation, feature embedding, self-supervised pretraining, and fine-tuning with linear probing, to facilitate a structured approach for feature learning and downstream task adaptation.

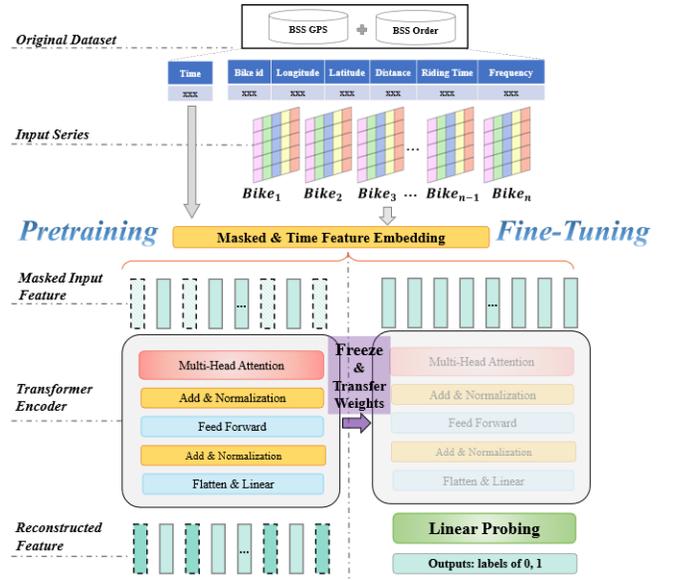

Fig. 5. The overall systematic solution pipeline

### C. Transformer with Self-supervised Pretraining

The proposed framework employs a Transformer architecture augmented with self-supervised pretraining to learn robust representations from unlabeled data. Departing from conventional Transformers that emphasize temporal coherence optimization, this proposed self-supervised Transformer (SSTransformer) incorporates multidimensional reconstruction objectives to enhance feature learning for identifying abnormal patterns. This section gives a thorough explanation of the core methodologies, encompassing the multi-head attention mechanism, self-supervised pretraining, fine-tuning via linear probing, and the design of the loss functions in different phases.

#### 1) Multi-Head Attention

The Transformer backbone utilizes multi-head attention mechanisms [19] to capture both local and global dependencies within sequential or structured inputs (e.g., time-series sensor data or spatial-temporal trajectories). The core idea is to enable the model to adaptively focus on the key information in the input sequence when generating the output sequence. The Scaled Dot-Product Attention used by Transformer achieves this through the interaction of the Query ($Q$), Key ($K$), and Value ($V$) vectors, which is given by the equation:

$$\text{Attention}(Q, K, V) = soft\,max\left(\frac{QK^T}{\sqrt{d_k}}\right)V \quad (3)$$

where $d_k$ is the dimensionality of the key vector $K$. $Q$, $K$, and $V$ are obtained through linear transformations of the inputs.

Multi-Head Attention (MHA) extends this concept by partitioning the $Q$, $K$, and $V$ into $H$ independent subspaces (called "heads"), each computing attention in parallel. The final outputs are then concatenated and linearly transformed:

$$\text{MHAttn}(Q, K, V) = Concat(head_1, \cdots, head_H)W^O \quad (4)$$

where $head_i = Attention(QW_i^Q, KW_i^K, VW_i^V)$, and $W_i^Q, W_i^K, W_i^V$, and $W^O$ are learnable parameters.

*2) Self-supervised pretraining*

Self-supervised learning (SSL) is a paradigm in ML where models learn meaningful representations from unlabeled data by generating supervision signals directly from the data itself. The proposed framework replaces traditional temporal embeddings with learnable embeddings, enabling the Transformer to intrinsically capture gradual feature transition dynamics and latent anomalous trend patterns (as empirically demonstrated in Fig. 3). This adaptive approach overcomes the rigid representations of static temporal encoding methods.

After feature embedding, a masked reconstruction task is utilized, where the model predicts and reconstructs randomly occluded segments of the input data features. Thus, the Transformer Encoder's primary objective is to minimize the Mean Absolute Error (MAE) between the original input $X$ and the reconstructed output. This task forces the model to infer contextual patterns and latent correlations, thereby facilitating feature learning and distilling domain-invariant features.

*3) Fine-Tuning with Linear Probing*

Following pretraining, the Transformer model is transferred and adapted to the downstream task of binary classification through linear probing. During this phase, all pre-trained Encoder layers are frozen with their weights transferred, and only a newly added linear head is trained. Unlike conventional Transformers that rely on full-parameter fine-tuning, the linear probing approach ensures a balance between task adaptation and retention of pre-trained knowledge, effectively reducing overfitting, particularly in low faulty sample scenarios.

*4) Loss Function*

In the pretraining phase, the Mean Absolute Error (MAE) is employed to quantify the discrepancy between the reconstructed data $x^r_{n,t,d}$ and the original input $x_{n,t,d}$. The MAE loss is computed as:

$$\mathcal{L}_{MAE} = \frac{1}{N*T*D} \sum_{n=1}^{N} \sum_{t=1}^{T} \sum_{d=1}^{D} |x_{n,t,d} - x^r_{n,t,d}| \quad (5)$$

During the fine-tuning phase, cross-entropy loss is used to evaluate the mismatch between predicted class probabilities and ground-truth labels. The cross-entropy loss is defined as:

$$\mathcal{L}_{CE} = -\frac{1}{N} \sum_{n=1}^{N} y_n \log \widehat{y_n} \quad (6)$$

where $y_n$ represents the ground-truth class label and $\widehat{y_n}$ denotes the predicted class probability for the $n$-th bike sample.

## IV. EXPERIMENTS AND RESULTS COMPARISON

### A. Data Description and Processing

This study employs real-world operational bike data comprising 10,730 instances (8,860 normal vs. 1,870 faulty vehicles), partitioned into an 80% training set (7,085 normal and 1,499 faulty samples) and a 20% test set (1,775 normal and 371 faulty samples) to ensure representative sampling. As detailed in Section 2.4, five discriminative features were selected as fault detection variables. The five features capture both static characteristics (e.g., prolonged idleness) and dynamic behavioral patterns (e.g., trajectory, frequent short-term usage) of bikes. This combination of static and dynamic data allows for a comprehensive examination of bike failure features using multidimensional analysis. To address heterogeneity in GPS sampling frequencies across bikes, all GPS trajectories were normalized to a uniform length based on the maximum sample count observed in the training set, thereby standardizing temporal resolution for model input.

### B. Baseline Models

This study primarily selects three categories of benchmarks:

- **Traditional ML models:** Decision Tree (DT), K-Nearest Neighbors (KNN), Support Vector Machine (SVM);
- **Ensemble ML models:** Random Forest (RF), XGBoost;
- **DL model:** Multilayer Perceptron (MLP), Gated Recurrent Unit (GRU), Long Short-Term Memory (LSTM), CNN+LSTM, Informer [20], and the Transformer without self-supervised pretraining.

### C. Evaluation Metric

To comprehensively assess model performance, this study selects Accuracy, Precision, Recall, and F1-score as metrics, which are computed based on True Positives (TP), False Positives (FP), True Negatives (TN), and False Negatives (FN).

Accuracy measures the overall correctness by calculating the ratio of correctly classified instances to the total instances, mathematically expressed as:

$$Accuracy = \frac{TP+TN}{TP+TN+FP+FN}. \quad (7)$$

Precision indicates the model's exactness in fault identification, with its formulation being:

$$Precision = \frac{TP}{TP+FP}. \quad (8)$$

Recall measures fault detection completeness, defined by:

$$Recall = \frac{TP}{TP+FN}. \quad (9)$$

F1-score represents the harmonic balance between Precision and Recall, calculated as:

$$F1\text{-}score = 2 \times \frac{Precision \times Recall}{Precision+Recall}. \quad (10)$$

Moreover, the model parameter size, represented as Params (M), along with the multiply-accumulate operations, denoted as MACs (G), serve as indicators of the DL models' complexity. The two metrics are frequently utilized to estimate models' computational complexity and real-time capabilities.

### D. Results Comparison

Table I presents the comparative performance metrics of the tested models. Several key insights emerge from these results, demonstrating the models' respective strengths and limitations. Ensemble ML methods, specifically RF and XGBoost, exhibit superior performance across accuracy, recall, precision, and F1-score compared to traditional ML models such as DT, KNN, and SVM. Notably, SVM achieves an exceptional recall of 0.99, indicating its enhanced specificity in identifying normal bikes. However, counterintuitively, baseline DL architectures, including MLP,

GRU, LSTM, and LSTM+CNN, demonstrate only moderate performance, failing to surpass the ensemble methods. This suggests that conventional DL architectures offer limited advantages in this context without customization, underscoring the need for more advanced and customized DL approaches tailored to the specific challenges of BSS fault detection.

The proposed self-supervised Transformer (SSTransformer) achieves the highest performance, with an accuracy of 0.9781, precision of 0.8889, and F1-score of 0.9358, outperforming all baseline models, including the same Transformer model without self-supervised pretraining (demonstrated in Table I). Despite higher computational complexity (3.59 M parameters, 3.09 G MACs), the SSTransformer's integration of multi-head attention mechanisms and self-supervised learning, together with the usage of selected five key ST features, proves highly effective for unusable bike detection in BSS.

TABLE I. PERFORMANCE COMPARISON OF THE TEST MODELS

| Model | ACC | Recall | Precision | F1 Score | MACs (G) | Params (M) |
|---|---|---|---|---|---|---|
| DT | 0.9455 | 0.8223 | 0.8733 | 0.8471 | --- | --- |
| KNN | 0.9478 | 0.8960 | 0.7898 | 0.8395 | --- | --- |
| SVM | 0.9646 | **0.9900** | 0.8032 | 0.869 | --- | --- |
| RF | 0.9734 | 0.9645 | 0.8787 | 0.9196 | --- | --- |
| XGBoost | 0.9730 | 0.9876 | 0.8518 | 0.9159 | --- | --- |
| MLP | 0.9618 | 0.9800 | 0.7925 | 0.8763 | 3.0236 | 3.5779 |
| GRU | 0.9719 | 0.9814 | 0.8518 | 0.9120 | 4.6827 | 3.4587 |
| LSTM | 0.9623 | 0.9865 | 0.7898 | 0.8772 | 4.5626 | 3.4237 |
| CNN+LSTM | 0.9724 | 0.9726 | 0.8625 | 0.9143 | 4.7134 | 3.5952 |
| Transformer | 0.9775 | 0.9879 | 0.8787 | 0.9301 | 3.0781 | 3.5830 |
| Informer | 0.9733 | 0.9728 | 0.8679 | 0.9174 | 3.0215 | 3.4653 |
| SSTransformer | **0.9781** | 0.9880 | **0.8889** | **0.9358** | 3.0853 | 3.5892 |

## V. CONCLUSION

The widespread adoption of bike-sharing systems (BSS) in urban environments has significantly improved commuter convenience but has also introduced escalating operational maintenance challenges. A critical yet underexplored research problem is the development of systematic methodologies to distinguish faulty unusable bikes from normal operational ones using historical data. To address this gap, this study proposes a novel self-supervised Transformer framework (SSTransformer) that leverages multi-dimensional ST features. Through a systematic analysis of real-world GPS trajectory data and trip records from an urban BSS, this study first identifies key discriminative ST signatures that differentiate faulty and operational bikes. Building on these insights and the identified key features, this study conducts comprehensive comparative experiments using authentic city-scale operational datasets to evaluate the proposed model against conventional ML, ensemble ML, and representative DL based baselines. The experimental results demonstrate that the proposed SSTransformer achieves superior fault detection performance, outperforming all the tested benchmarks. These findings validate the framework's potential as an innovative and scalable solution for intelligent maintenance in shared mobility systems.